\pdfoutput=1

\documentclass[11pt]{article}
\usepackage{kotex}

\usepackage[]{ACL2023}

\usepackage{times}
\usepackage{latexsym}
\usepackage{adjustbox}
\usepackage{graphicx}
\usepackage{amsmath}
\usepackage{multicol}

\usepackage[T1]{fontenc}

\usepackage[utf8]{inputenc}

\usepackage{microtype}

\usepackage {inconsolata}
\usepackage{setspace}
\usepackage{tabularx,ragged2e}
\usepackage{algorithm}
\usepackage{algpseudocode}
\usepackage{amsmath}
\usepackage{graphicx}
\usepackage{array}
\usepackage{adjustbox}
\usepackage{tabularray}
\usepackage{boldline}
\usepackage{multirow}
\usepackage{subcaption}
\usepackage{xcolor}
\usepackage{forest}
\usepackage{caption}
\usepackage{lipsum}
\newcolumntype{C}{>{\Centering\arraybackslash}X} 

\title{Class Granularity: How richly does your knowledge graph represent the real world?}

\author{Sumin Seo, Heeseon Cheon, Hyunho Kim\\
  \texttt{sumin.seo@navercorp.com}, \texttt{heeseon.cheon@navercorp.com},\\ \texttt{kim.hh@navercorp.com}}

\begin{document}
\maketitle
\begin{abstract}
To effectively manage and utilize knowledge graphs, it is crucial to have metrics that can assess the quality of knowledge graphs from various perspectives. While there have been studies on knowledge graph quality metrics, there has been a lack of research on metrics that measure how richly ontologies, which form the backbone of knowledge graphs, are defined or the impact of richly defined ontologies. In this study, we propose a new metric called \textit{Class Granularity}, which measures how well a knowledge graph is structured in terms of how finely classes with unique characteristics are defined. Furthermore, this research presents potential impact of \textit{Class Granularity} in knowledge graph's on downstream tasks. In particular, we explore its influence on graph embedding and provide experimental results. Additionally, this research goes beyond traditional Linked Open Data comparison studies, which mainly focus on factors like scale and class distribution, by using \textit{Class Granularity} to compare four different LOD sources.
\end{abstract}

\section{Introduction}

A knowledge graph is a data system comprising RDF triples structured as \textit{`subject-predicate-object'}, enabling it to represent real-world entities and their associated relationships. Within the domain of information retrieval \cite{reinanda2020knowledge} and question-answering \cite{lan2021survey} tasks, knowledge graphs have traditionally assumed a prominent role. More recently, their significance has grown within the field of AI. They are integrated into language models as external repositories of knowledge, thereby augmenting the precision and reasoning capabilities of these models \cite{schneider2022decade}. Furthermore, knowledge graphs are harnessed to enrich content-related information within recommendation systems \cite{shao2021survey}.

However, the efficacy of knowledge graphs relies on their quality. Consequently, various research efforts have been undertaken to measure the quality of knowledge graphs. In particular, it is essential to consider ontology, which plays a foundational role as the backbone of knowledge graphs. An ontology is a schema that formally declares classes, which represent types of instances, along with the predicates associated with each class and the hierarchical relationships between classes and predicates. \citet{gruber1993ontology} defines ontology as "explicit specifications of conceptualizations". The comparison made in \citet{raad2015survey} between a machine's knowledge graph and a human's level of knowledge suggests that both are crucial for decision-making. Just as humans rely on their knowledge to make decisions alongside their reasoning abilities, machines also heavily depend on their knowledge graph, or ontology, to enhance their reasoning capacity. For instance, when someone encounters the names \textit{"Let it be"} and \textit{"Hey Jude"}, varying levels of comprehension may be observed among different individuals. Some individuals may fail to discern their meaning, while others may categorize them as "songs". Furthermore, certain people might even recognize that both songs share the same artist,\textit{The Beatles}.

There are various methods available for measuring the quality of ontology and knowledge graphs. However, in this study, we aim to assess how granular a knowledge graph is using the \textit{Class Granularity} metric. The granularity of a knowledge graph, in essence, refers to the presence of numerous defined predicates and the high depth and breadth of the ontology. Despite its significance, there has been a lack of research into metrics for measuring granularity and its impact. According to \citet{10.1007/978-3-642-10871-6_5}, "in general, richly populated ontologies, with higher depth and breadth variance are more likely to provide reliable semantic content". Furthermore, in this study, we analyze the impact of granularity through graph embedding techniques and knowledge base question answering (KBQA). Additionally, we provide comparative analysis results on the level of granularity in Linked Open Data sources such as YAGO, DBpedia, and Wikidata, which, to our knowledge, have not been previously reported.

The contributions of this study are as follows:

\begin{itemize}
\item We propose a metric that takes into account both the ontology and the knowledge graph instances for measuring the granularity of knowledge graphs.
\item We provide the first comparative results on the level of granularity in Linked Open Data sources.
\item We conduct experiments to assess the impact of granularity on specific tasks.
\end{itemize}

\section{Terminology}
\subsection{Class}
\textit{Class} is a "concept in a domain of discourse" \cite{gruber1993ontology}. Sometimes it is called \textit{Concept} or \textit{Type}. For example, \textit{Person}, \textit{Movie} and \textit{Nation} can be classes of open-domain knowledge graph. 

In ontology, a hierarchy structure is established between classes. This hierarchy is typically defined in the form of \textit{"A - is subclass of - B"},  where \textit{A} represents the subclass and \textit{B} represents the superclass. For example, in ontology, relationship like \textit{"Athlete - is subclass of - Person"} can be defined where \textit{Athlete} is the subclass and \textit{Person} is the superclass. The classes that shares the same direct superclass are sibling classes. For instance, if \textit{"Athlete - is subclass of - Person"} and \textit{"Politician - is subclass of - Person"}, \textit{Athlete} and \textit{Politician} are sibling classes. The superclass that does not have any parent classes above it is referred to as the root class. In the context of Linked Open Data, this root class is often represented using classes like \textit{Entity} or \textit{Thing}. 

\subsection{Predicate}
\textit{Predicate} is a "property of each concept describing various features and attributes of the concept" \cite{gruber1993ontology}. It is also called \textit{Property} or \textit{Attribute}. 

Predicates are associated with each class are defined in the ontology. When there exists a subclass-superclass relationship, the predicates used in the superclass can also be used by the subclass. For instance, if \textit{Facility} is the superclass of \textit{Museum}, and \textit{Facility} has predicates like \textit{longitude} and \textit{latitude}, then \textit{Museum} can also use these properties. Additionally, as a subclass, \textit{Museum} may have additional properties, such as \textit{collection or exhibition size}, that are specific to its classification.

However, in real-world knowledge graphs, it's common to find triples that don't strictly adhere to the predicates defined in the ontology. In other words, instances of a class may have triples with properties that are not explicitly defined for that class in the ontology. This flexibility allows knowledge graphs to capture a broader range of information, even if they deviate from the strict definitions in the ontology.

\subsection{Instance}
\textit{Instance} is an entity that can be a subject or object of the triple that consists knowledge graph. It is also called \textit{Entity} or \textit{Individual}. For example, \textit{"Mission Impossible"} is an instance and its class is \textit{Movie}. \textit{"Mission Impossible"} can be a subject of the triple like \textit{"Mission Impossible - cast member - Tom Cruise"} and can be an object of the triple like \textit{"Ethan Hunt - character of - Mission Impossible"}.

\section{Previous Works}

Ontology quality evaluation and Linked Data quality evaluation indeed share common ground in terms of assessing knowledge graph quality. However, they differ in specific metrics and perspectives. One commonality between them is the need for more comprehensive approaches to measure the granularity of data.

While ontology quality evaluation typically focuses on assessing the structure, relationships, and consistency of the ontology itself, Linked Data quality evaluation often concentrates on the quality of data instances and the links between them. Consequently, the specific metrics and criteria they use may differ.

\begin{table*}[htbp]
    \centering
    \caption{Linked Data Quality dimensions}
    \label{tab:linked data quality dimensions}
    \begin{tabular}{|c|p{10cm}|} 
        \hline
        Accessibility dimensions & availability, licensing, interlinking, security, performance \\
        \hline
        Intrinsic dimensions & syntactic validity, semantic accuracy, consistency, conciseness, completeness \\
        \hline
        Contextual dimensions & relevancy, trustworthiness, understandability, timeliness \\
        \hline
        Representational dimensions & representational conciseness, interoperability, interpretability, versatility \\
        \hline
    \end{tabular}
\end{table*}

\subsection{Ontology Quality Evaluation}
In prior research on ontology quality evaluation, four distinct approaches can be delineated \cite{lourdusamy2018review,brank2005survey,raad2015survey}. \textit{Gold Standard Evaluation} measures how closely an ontology aligns with a high-quality benchmark ontology when such an ontology is available. \textit{Data-Driven Evaluation} assesses the extent to which an ontology effectively encapsulates domain-specific information by methods like keyword extraction from the domain's corpus. \textit{Application or Task-Based Evaluation} examines performance in downstream tasks. \textit{Metric-Based Evaluation}, also known as structure-based evaluation, employs formula that discern the structure and statistical characteristics of classes and predicates to gauge quality. This study aligns with Metric-Based Evaluation.

In structural evaluation, various metrics are employed to compare knowledge graphs. These metrics encompass the "schema metric," the "instance/knowledgebase metric," the "class metric," the "graph metric," and the "complexity metric." \cite{lourdusamy2018review,tartir2005ontoqa}, The \textit{Schema Metric} assesses the number of classes, properties, and properties per class, with a particular emphasis on the ontology. The \textit{Instance/Knowledgebase Metric} involves calculating statistics such as the average number of instances per class, taking into account instances within the context of an ontology. In the case of the \textit{Class Metric}, it involves calculating either the number of instances of the "Person" class or the degree of connection of the "Person" class with other classes, thereby representing the characteristics of each class. The \textit{Graph metric} entails the application of fundamental graph theory statistics, including cohesion and cardinality.

There were metrics designed to measure the granularity of knowledge graphs, but they had limitations. OntoQA\cite{tartir2005ontoqa} introduces \textit{Relationship Richness}, \textit{Attribue Richness} and \textit{Inheritance Richiness} as \textit{Schema Metrics} to evaluate ontology's richness. \textit{Relationship Richness} measures the proportion of predicates that are relationships (e.g."subclass of") relative to the total number of predicates. \textit{Attribute Richness} calculates the average number of predicates defined per class. \textit{Inheritance Richness} quantifies the average number of subclasses per class. These metrics do not simultaneously consider both predicates and classes. For example, a high \textit{Inheritance Richness} score may indicate a high level of ontology granularity, but it does not necessarily imply meaningful granularity unless subclasses have predicates distinct from their superclasses. For instance, even if \textit{Athlete} is defined as a subclass of \textit{Person},  if instances of \textit{Athlete} share the same predicates as \textit{Person} and do not introduce additional predicates like \textit{sports}, a high Inheritance Richness score alone may not adequately signify effective granularity.

The \textit{Density Measure} presented in the \citet{alani2006metrics} can also be used to assess the granularity of an ontology. This metric involves calculating the weighted sum of predicates, superclasses, subclasses, and sibling classes for each class in the ontology, and then averaging these values. While it does take into account both classes and predicates, it doesn't provide information about whether the fine-grained classes or predicates are actually used in instances. Therefore, it cannot determine whether the knowledge graph effectively reflects a finely-grained ontology. \textit{Class Granularity} measures how many classes have added predicates that are not defined for superclass, and whether these added predicates are employed within instances.

Previous works suggest important dimensions to consider when evaluating ontologies: \textit{Accuracy}, \textit{Completeness}, \textit{Conciseness}, \textit{Adaptability}, \textit{Clarity}, \textit{Computational efficiency} and \textit{Consistency} \cite{vrandevcic2009ontology,obrst2007evaluation,gomez2004ontology,gangemi2006modelling}. High \textit{Class Granularity} is related to high \textit{Completeness} as \textit{Completeness} "measures if the domain of interest is appropriately covered in this ontology \cite{raad2015survey}". However, in the context of \textit{Adaptability}, which "measures how far the ontology
anticipates its uses \cite{raad2015survey}", high \textit{Class Granularity} can have a negative impact. This becomes particularly relevant when the schema is designed for future utilization but unused in instances yet. Regarding four quality categories (\textit{Content}, \textit{Language}, \textit{Methodology}, \textit{Tools}) of OntoMetric\cite{lozano2004ontometric}, \textit{Class Granularity} is falls under the category of \textit{Content}("content of ontology, organization of their contents").

\subsection{Linked Data Quality Evaluation}
Linked Open Data is a concept related to making data freely available and easily accessible on the internet in a structured and interlinked manner. \textit{Wikidata}, \textit{DBpedia}, \textit{YAGO}, and \textit{Freebase} are well-known examples of Linked Open Data (LOD) sources. Many studies comparing the quality of Linked Open Data (LOD) primarily focus on comparing the scale of triples, the distribution of classes and predicates within the data, the depth of ontologies, and the incoming/outgoing degree of data entities.\cite{ringler2017one,farber2018knowledge,heist2020knowledge}

Furthermore, in \citet{zaveri2016quality}, four dimensions were introduced to assess the quality of Linked Data, namely the \textit{Accessibility dimensions}, \textit{Intrinsic dimensions}, \textit{Contextual dimensions} and \textit{Representational dimensions}. These dimensions were further categorized into subcategories as in Table ~\ref{tab:linked data quality dimensions}, with recommended metrics for their measurement. Other studies \cite{debattista2018evaluating,debattista2016luzzu,ringler2017one} applied the framework to conduct a comparative analysis of Linked Data. While this framework provides a much richer analysis compared to previous comparative studies, it does not include an analysis of the granularity of ontologies within each Linked Open Data (LOD) source and how well this granularity is reflected in the actual RDF data. Among the metrics introduced in \citet{zaveri2016quality}, IN3 is a metric within the \textit{Interpretability} dimension which quantifies the "invalid usage of undefined classes
and properties". It is similar to \textit{Class Granularity} in that it considers Linked Data's ontology and its instantiation. However, it cannot measure the level of granularity in the data. In this study, we provide \textit{Class Granularity} for \textit{Wikidata}, \textit{DBpedia}, \textit{YAGO}, and \textit{Freebase}, allowing us to compare the level of granularity in LOD (Linked Open Data) that has not been addressed in previous research.

\section{Class Granularity}
\subsection{Definition}
Class Granularity is a metric that can measure how detailed the ontology of a knowledge graph is and how well it reflects the actual knowledge graph composed of RDF triples. In this section, the definition of the \textit{Instance with Distinct Predicate Proportion Average} for calculating \textit{Class Granularity} is first provided, followed by the introduction of the formula for \text{Class Granularity}.

\subsubsection{Instance with Distinct Predicate Proportion Average}\mbox{}\\
\textit{Instance with Distinct Predicate Proportion Average (IDPPA)} is the average of \textit{Instance with Distinct Predicate Proportion}, so \textit{Distinct Predicate} and \textit{Instance with Distinct predicate proportion (IDPP)} should be defined first. We define \textit{distinct predicates} of a specific class as predicates that do not exist in the ontology-defined superclass or sibling classes but are present in the specific class. For example, if \textit{Athlete} is a subclass of \textit{Person} and \textit{Politician} and \textit{Artist} are sibling classes, and \text{Athlete} has its unique predicates like \textit{sport} and \textit{position played on the team} that are not found in \textit{Person}, \textit{Politician}, or \textit{Artist}, then these predicates are considered as \textit{distinct predicates}. \textit{Distinct predicates} serve the role of adding unique characteristics to a class. \textit{IDPP} of a specific \textit{distinct predicate} within a particular class is the proportion of instances that possess the predicate in the total instances in that class.\\
Each class's \textit{IDPPA} represents the average of \textit{IDPP} for its \textit{distinct predicate}. For instance, if the \textit{Athlete} class has two \text{distinct predicates}, namely \textit{sport} and \textit{position played on the team}, and if 90\% of \textit{Athlete} instances have the predicate \textit{sport} (e.g., \textit{Michael Jordan - sport - basketball}), and 50\% have the predicate \textit{position played on the team} (e.g., \textit{Lionel Messi - position played on the team - forward}), then the \textit{IDPPA} for \textit{Athlete} is calculated as \begin{math}(0.9 + 0.5) / 2 = 0.7\end{math}.\\
However, to prevent the indiscriminate addition of duplicate predicates to all classes, a penalty is applied in the calculation of \textit{IDPPA}. \textit{Distinct predicates} should be predicates that differentiate the class from others. Therefore, for instance, defining \textit{birth place} under \textit{Athlete} rather than its superclass \textit{Person}, or defining \textit{member of political party} under \textit{Athlete} instead of its sibling class \textit{Politician}, may not be desirable. Classes should have distinct characteristics from their superclass and sibling classes. If a superclass or sibling class of a certain class has a higher proportion of instances with \textit{distinct predicates} than that class itself, the IDPP is treated as 0. This is because even though \textit{distinct predicates} are defined for that class, they are actually more prominently associated with superclass or sibling classes, and they do not serve as distinguishing features for that class. Expanding on the previous example, if the ratio of \textit{position played on the team} in the \textit{Person} class is 0.8, which is higher than the 0.5 in \textit{Athlete}, it may not represent a unique characteristic of \textit{Athlete}. In this case, the \textit{IDPPA} for \textit{Athlete} would be calculated as the average of 0.9 for \textit{sport} and 0 for \textit{position played on the team}, resulting in \begin{math}(0.9 + 0) / 2 = 0.45\end{math}.

Finally, equations ~\ref{IDPP ratio formula} and ~\ref{IDPPA ratio formula} represent the formulas for calculating IDPPA. In equation ~\ref{IDPP ratio formula}, \textit{IPP} stands for \textit{Instance with Predicate Proportion}, where \begin{math}IPP(P|C)\end{math} is the ratio of instances in a specific class \textit{C} that have a particular predicate \textit{P} when class \textit{C} is given. In equation ~\ref{IDPP ratio formula}, \textit{DP} stands for \textit{Distinct Predicate} of class \textit{C}, and \textit{RC} represents \textit{Related Class}, which includes \textit{C}'s superclass and sibling classes. \begin{math}IPR(DP|RC)_{max}\end{math} is the value when among \textit{C}'s related classes, the ratio of instances containing \textit{DP} is the highest. In equation ~\ref{IDPPA ratio formula}, \begin{math}N_{dp}\end{math} denotes the number of distinct predicates for class \textit{C}, and \begin{math}dp_i\end{math} represents the i-th distinct predicate.

\begin{equation}
\label{IDPP ratio formula}
\hspace*{0.7cm}
\text{\small IDPP(DP | C) =} 
\begin{cases}
    \text{\small IPP(DP|C)} & \\
    \text{\tiny  when $IPP(DP|RC)_{max} \leq IPR(DP|C)$}  \\
    \text{\small 0} & \\
    \text{\tiny when $IPP(DP|RC)_{max} > IPR(DP|C)$} \\
\end{cases}
\hspace*{0.7cm}
\end{equation}

\begin{equation} 
\label{IDPPA ratio formula}
\hspace*{0.7cm} 
\text{ IDPPA(C) =}
\begin{cases}
    \text{ $\frac{\sum_{i = 1}^{N_{dp}} \text{IDPP(dp}_i | C\text{)}}{N_{dp}}$} & \\
    \text{\small when $N_{dp} > 0 $}  \\
    \text{ 0} & \\
    \text{\small when $N_{dp} = 0$} \\
\end{cases}
\hspace*{1cm} 
\end{equation}

\subsubsection{Class Granularity}\mbox{}\\
\textit{Class Granularity} is the average of \textit{IDPPA} calculated for all classes except the root class in the ontology. In cases where only the root class exists, \textit{Class Granularity} is 0 because it signifies a complete lack of granularity or subclassification in the ontology. Equation ~\ref{Class Granularity ratio formula} represents the formula for calculating Class Granularity. \begin{math}N_{c}\end{math} signifies the total number of classes in the ontology, and when \begin{math}N_{c}=1\end{math}, it indicates a scenario where only the root class exists. \begin{math}c_{i}\end{math} represents the i-th class in the ontology.

\begin{equation}
\label{Class Granularity ratio formula}
\hspace*{0.7cm}
\text{\small Class Granularity} =
\begin{cases}
    \frac{\sum_{i = 1}^{N_{c}-1}\text{\small IDPPA(c}_i\text{\small)}}{N_{c}-1} & \\
    \text{\scriptsize when $N_{c} > 1$}  \\
    \text{\small 0} & \\
    \text{\scriptsize when $N_{c} = 1$} 
\end{cases}
\hspace*{0.7cm}
\end{equation}

Algorithm 1 is the pseudo code outlining the process for calculating \textit{Class Granularity} when provided with a knowledge graph and its ontology. \textit{all\_class\_list} in line 4 means a list of all the classes defined in the ontology. \textit{class.all\_predicate\_list} in line 9 represents the list of all predicates used in RDF triples where the instances of the class are the subjects, regardless of ontology definitions. For example, the ratio of instances that has \textit{member of political party} in \textit{Athlete} class is calculated even if \textit{member of political party} is not defined for \textit{Athlete} in the ontology. On the other hand, \textit{class.distinct\_predicate\_list} in line 16 is a list that includes only predicates defined in the ontology.

\begin{algorithm*}
\caption{Calculate Class Granularity}
\small 
\begin{algorithmic}[1]
\State Initialize an empty dictionary $\text{IPR\_dic}$
\State Initialize a variable $\text{total\_IDPP} \gets 0$
\State Initialize a variable $\text{class\_count} \gets 0$
\If{$\text{length of } \text{all\_class\_list} = 1$}
    \State Set $\text{Class Granularity} \gets 0$
\Else
    \For{each $\text{class}$ in $\text{all\_class\_list} - [\text{root\_class}]$}
        \State Create an empty dictionary $\text{class\_instance\_ratio\_dic}$
        \For{each $\text{predicate}$ in $\text{class.all\_predicate\_list}$}
            \State Calculate and store the instance ratio with predicate in $\text{class\_instance\_ratio\_dic}$ with $\text{predicate}$ as the key
        \EndFor
        \State Store $\text{class\_instance\_ratio\_dic}$ in $\text{IPR\_dic}$ with $\text{class}$ as the key
    \EndFor
    \For{each $\text{class}$ in $\text{all\_class\_list} - [\text{root\_class}]$}
        \State Initialize $\text{class\_IDPP} \gets 0$
        \For{each $\text{distinct\_predicate}$ in $\text{class.distinct\_predicate\_list}$}
            \State Calculate $\text{max\_related\_class\_IPR} \gets \max(\text{IPR\_dic[related\_class][distinct\_predicate]}$ for each $\text{related\_class}$ in $\text{related\_class\_list})$
            \If{$\text{IPR\_dic[class][distinct\_predicate]} > \text{max\_related\_class\_IPR}$}
                \State $\text{predicate\_IDPP} \gets \text{IPR\_dic[class][distinct\_predicate]}$
            \Else
                \State $\text{predicate\_IDPP} \gets 0$
            \EndIf
            \State Add $\text{predicate\_IDPP}$ to $\text{class\_IDPP}$
        \EndFor
        \State Add $\text{class\_IDPP}$ to $\text{total\_IDPP}$
        \State Increment $\text{class\_count}$ by $1$
    \EndFor
    \State Calculate the average IDPP as $\text{average\_IDPP} \gets \frac{\text{total\_IDPP}}{\text{class\_count}}$
    \State $\text{Class Granularity} \gets \text{average\_IDPP}$
\EndIf
\State \textbf{Output:} $\text{Class Granularity}$
\end{algorithmic}
\end{algorithm*}

\subsection{Calculation Example}\label{Calculation Example}

This section demonstrates through examples that as the ontology becomes more finely defined, \textit{Class Granularity} increases. Consider a knowledge graph consisting of 5 instances and 9 triples, as shown in Table ~\ref{tab:Class Granularity Calculation Example Triples} (The IDs like "Q714" and "P161" within the parentheses are actual identifiers used in Wikidata.). In this knowledge graph, \textit{Class Granularity} can be calculated for both a "less granularity ontology (ontology A)" with 3 classes and a "more granularity ontology (ontology B)" with 9 classes which is depicted in Table ~\ref{tab:Class Granularity Calculation Example Ontologies}. Both ontologies have \textit{Creative Work} as the root class. Additionally, when a class has subclasses, the instances of that class are considered to include instances from its subclasses. For instance, the instances of \textit{Creative Work} encompass both the instances of \textit{Audio Work} and \textit{Visual Work}.

The \textit{Class Granularity} for the "less granularity ontology" is calculated as follows: First, calculate the IDPPA for class \textit{Audio Work} and class \textit{Visual Work}, excluding the root class. For the \textit{distinct predicates} of class \textit{Audio Work}, compare the \textit{IPR (Instance with predicate proportion)} of each predicate with the \textit{IPRs} of the related classes — class \textit{Creative Work} (a superclass) and class \textit{Visual Work} (a sibling class). For predicate \textit{performer}, class \textit{Audio Work}'s \textit{IPR} is $\frac{2}{2}$ (as both \textit{"News of the World"} and \textit{"Isn't She Lovely"} have this predicate), class \textit{Creative Work}'s \textit{IPR} is $\frac{2}{5}$ , and class \textit{Visual Work}'s \textit{IPR} is $\frac{0}{3}$ . As class \textit{Audio Work}'s \textit{IPR} , $\frac{2}{2}$ is greater than maximum IPR among related classes, the IDPP for predicate \textit{performer} is $\frac{2}{2}$ = 1. Following a similar calculation, predicate \textit{tracklist}'s \textit{IDPP} is $\frac{1}{2}$ (as only \textit{"News of the World"} has this predicate) and predicate \textit{lyrics by}'s is $\frac{1}{2}$ (as only \textit{"Isn't She Lovely"} has this predicate). Thus, the IDPPA for class \textit{Audio Work} is $(\frac{2}{2} + \frac{1}{2} + \frac{1}{2}) / 3 \approx 0.67$. Caluation for the "more granularity ontology"'s each \textit{IDPPA} is shown in Table ~\ref{tab:Class Granularity Calculation Example Ontologies} and its \textit{Class Granularity} is $(\frac{1}{1} + \frac{1}{1} + \frac{1}{1} + 0 + \frac{1}{1} + \frac{1}{1} + \frac{1}{1} + \frac{1}{1}) / 8 = 0.875$.

As demonstrated in the examples, \textit{Class Granularity} increases as the ontology becomes is constructed more richly with an abundance of classes and predicates in a meaningful manner for the same knowledge graph. However, simply adding classes and predicates to the ontology without careful consideration does not necessarily lead to higher \text{Class Granularity}. For instance, let's consider the "more granularity ontology" where we add two classes, \textit{Horror Movie} and \textit{Action Movie} as subclasses under \textit{Movie}. In this case, there are no instances that belong to these two classes in the knowledge graph(suppose \textit{"Titanic"} does not belong to any of them), resulting in an \textit{IDPPA} of 0 for each class and a decrease in Class Granularity to 0.7 ($(\frac{1}{1} + \frac{1}{1} + \frac{1}{1} + 0 + \frac{1}{1} + \frac{1}{1} + \frac{1}{1} + \frac{1}{1} + 0 + 0) / 10 = 0.7$). Additionally, if we add predicates like \textit{attendance}, which are not used in the class \textit{Movie} class in the knowledge graph, or duplicate predicates like \textit{cast member of} from the superclass \textit{Video}, it can also lead to a decrease in Movie's IDPP and, subsequently, a decrease in \textit{Class Granularity}. Therefore for high \textit{Class Granularity} it's essential to reflect the classes and predicates that are actually used in the knowledge graph and ensure that differences between classes are reflected through distinct predicates.

\begin{table*}[]
\centering
\caption{Example Triples for Class Granularity Calculation}
\label{tab:Class Granularity Calculation Example Triples}
\small
\begin{tabular}{|l|l|l|}
\hline
Instance          & RDF triples                                                                                                                                                                                & Class       \\ \hline
News of the World & \begin{tabular}[c]{@{}l@{}}•News of the World (Q309021) - performer (P175) - Queen (Q15862)\\ •News of the World (Q309021) - tracklist (P658) - We Will Rock You (Q66010300)\end{tabular}  & Music Album \\ \hline
Isn't She Lovely  & \begin{tabular}[c]{@{}l@{}}•Isn't She Lovely (Q2530274) - performer (P175) - Stevie Wonder (Q714)\\ •Isn't She Lovely (Q2530274) - lyrics by (P676) - Stevie Wonder (Q714)\end{tabular}    & Music Song  \\ \hline
The Starry Night  & •The Starry Night (Q45585) - collection (P195) - Museum of Modern Art (Q188740)                                                                                                            & Painting    \\ \hline
Titanic           & \begin{tabular}[c]{@{}l@{}}•Titanic (Q44578) - cast\_member (P161) - Leonardo DiCaprio (Q38111)\\ •Titanic (Q44578) - box office (P2142) - 1,850,197,130 United States dollar\end{tabular} & Movie       \\ \hline
Friends           & \begin{tabular}[c]{@{}l@{}}•Friends (Q79784) - cast\_member (P161) - Jennifer Aniston (Q32522)\\ •Friends (Q79784) - original broadcaster (P449) - NBC (Q13974)\end{tabular}               & TV Series   \\ \hline
\end{tabular}
\end{table*}

\subsection{Characteristics of the Metric}

This section examines the properties of \text{Class Granularity} as a data quality metric. According to \citet{heinrich2018requirements}, a data quality metric that possesses the properties of the "existence of minimum and maximum metric values" and "the interval scaling of metric values" is considered "inherently interpretable in terms of the measurement unit" and can be represented as a percentage. \textit{Class Granularity} meets both of these properties.

\subsubsection{Minimum and Maximum}\mbox{}\\

\textit{Class Granularity} has a minimum value of 0 in three cases: First, when only the root class is defined in the ontology. Second, when classes are defined and subdivided, but there are no instances of those classes in the knowledge graph. Third, classes are defined, instances exist, but distinct predicates are not used. In all of these cases, it is difficult to consider the ontology as well-reflecting and sufficiently subdividing the knowledge graph. The maximum value of \textit{Class Granularity} is 1. This occurs when, for every class defined in the ontology, all instances of that class possess the class's distinct predicates. 

\subsubsection{Interval Scale}\mbox{}\\
\textit{Class Granularity} exhibits the characteristics of an interval scale. According to \citet{allen2001introduction}, it is always possible to achieve a transformation between any two interval scales by implementing a positive linear change of the type $x \rightarrow ax + b$ (where $a > 0$). \textit{Class Granularity} represents the average ratio of how extensively distinct predicates are being used for each class. When Class Granularity increases from 0.5 to 0.6 and from 0.6 to 0.7, the interval retains the same meaning.

\begin{table*}
\centering
\caption{Example Ontologies for Class Granularity Calculation }
\label{tab:Class Granularity Calculation Example Ontologies}
\small
\begin{tblr}{
  colspec = {X[0.5] X[0.9] X[1.5] X[1.1] X},
  cell{2}{1} = {r=3}{},
  cell{5}{1} = {r=9}{},
  vlines,
  hline{1-2,5,14} = {1.5pt},
  hline{3-4,6-13} = {2-5}{},
  rowsep=0.9pt,
}
                 &            Class                          & Distinct Predicates & Instances (refer to Table ~\ref{tab:Class Granularity Calculation Example Triples})   & IDPPA \\
ontology A\newline  (less \newline granularity ontology) & Creative Work                        &                     &     &     \\
& \textbar{}\_\_ Audio Work            & performer, tracklist, lyrics by      & News of the World, Isn't She Lovely   & \tiny\(\left(\frac{2}{2} + \frac{1}{2} + \frac{1}{2}\right) / 3  \approx 0.67\) \\
                 & \textbar{}\_\_Visual Work & collection, cast member, box office, original broadcaster & The Starry Night, Titanic, Friends & \tiny\(\left( \frac{1}{3} + \frac{2}{3} + \frac{1}{3} + \frac{1}{3} \right) / 4  \approx 0.33\) \\
ontology B\newline  (more \newline granularity ontology)& Creative Work                        &                   &      &   \\
                 & \textbar{}\_\_ Audio Work            & performer                        & & \text{\normalsize \begin{math} (\frac{2}{2} ) / 1 = 1 \end{math}}\\
                 &  \hspace{1em} \textbar{}\_\_\_\_ Music Album       & tracklist           & News of the World   &  \text{\normalsize \begin{math} (\frac{1}{1} ) / 1 = 1 \end{math}}\\
                 &  \hspace{1em} \textbar{}\_\_\_\_ Music Song        & lyrics by                  & Isn't She Lovely     & \text{\normalsize \begin{math} (\frac{1}{1} ) / 1 = 1 \end{math}} \\
                 & \textbar{}\_\_ Visual Work           &                   &     &  0  \\
                 &  \hspace{1em}\textbar{}\_\_\_\_ Painting          & collection                  & The Starry Night    & \text{\normalsize \begin{math} (\frac{1}{1} ) / 1 = 1 \end{math}} \\
                 &  \hspace{1em}\textbar{}\_\_\_\_ Video             & cast member                  &   &  \text{\normalsize \begin{math} (\frac{2}{2} ) / 1 = 1 \end{math}} \\
                 &  \hspace{2em}\textbar{}\_\_\_\_\_\_\_\_ Movie     & box office                  & Titanic   &   \text{\normalsize \begin{math} (\frac{1}{1} ) / 1 = 1 \end{math}} \\
                 &  \hspace{2em}\textbar{}\_\_\_\_\_\_\_\_ TV Series & original broadcaster             & Friends     & \text{\normalsize \begin{math} (\frac{1}{1} ) / 1 = 1 \end{math}}
\end{tblr}
\end{table*}

\section{When Class Granularity Is Helpful?}
As seen in ~\ref{Calculation Example}, the higher the \textit{Class Granularity}, the more easily one can anticipate the structure of the knowledge graph just by examining the ontology. Additionally, a higher \textit{Class Granularity} implies that classes possess distinct characteristics and are well-subdivided. This section provides a more detailed examination of downstream tasks related to \textit{Class Granularity}.

\subsection{Graph Embedding}

There have been studies showing that incorporating rich ontology information can lead to improved performance in knowledge graph embeddings \cite{jain2021improving,li2023transo,diaz2018embeds}. This section aims to demonstrate through experiments that when an ontology is well-defined with granularity, important concepts can be embedded in a similar space during the utilization phase, which can be beneficial. For instance, if individuals with similar professions are intended to have similar embedding values, then subdividing the \textit{Person} class into various professions can achieve this goal.

\subsubsection{Experiment Settings}\mbox{}\\
RDF triples are obtained from Wikidata triples with human entities as subjects. An experimental knowledge graph is constructed  by extracting 2-hop relations. Then, comparison was made by assigning classes to entities based on three different ontologies, as shown in Table ~\ref{tab:example_ontologies}. For class assignments, \textit{Person} is assigned when an entity is an instance of \textit{human (Q5)} in Wikidata and \textit{Thing} otherwise. For the subdivided classes like \textit{Actor} and \textit{Athlete}, \textit{occupation (P106)} property is referred. Based on the class assignments, RDF triples are added in the form of \textit{"entity - instance of (P31) - class"} to the graph. For instance, entities that are instances of \textit{Actor} in Example3, become instances of \textit{Artist} in Example2 and become instances of \textit{Person} in Example3. TransE model \cite{bordes2013translating} provided by pykeen\footnote{https://github.com/pykeen/pykeen} 1.10.1 with default training parameters is used for graph embedding. 

\subsubsection{Graph Embedding Results}
\begin{figure*}[htbp]
  \centering
    \caption{t-SNE Visualization of Graph Embedding Examples}
      \vspace{10pt} 
    \includegraphics[width=\linewidth]{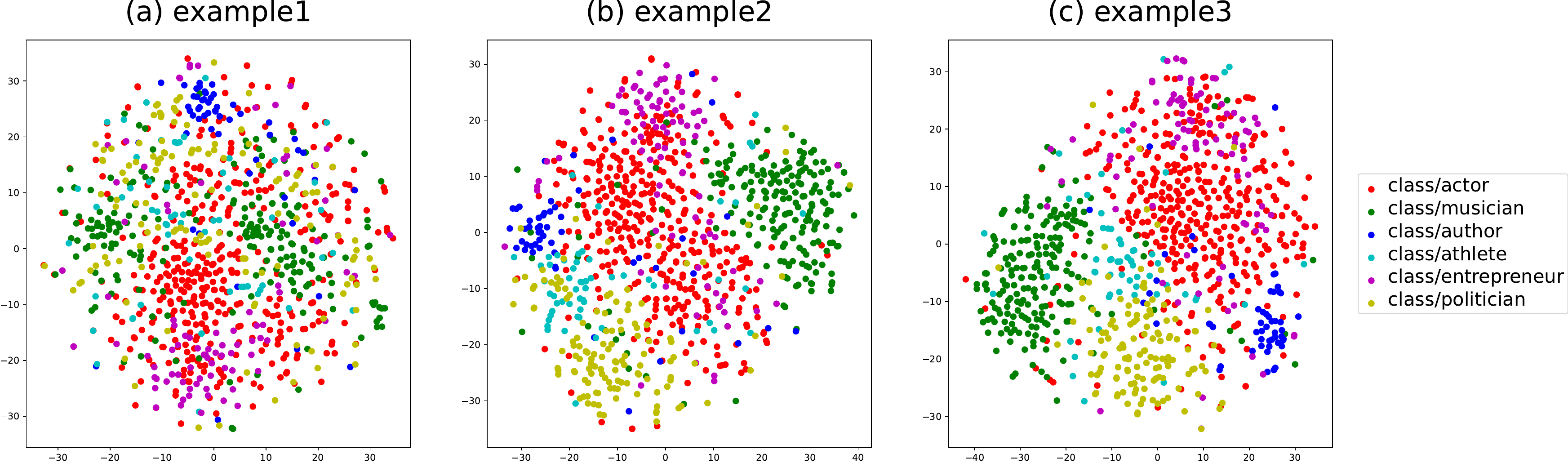}
  \label{fig:t-SNE Visualization of Graph Embedding Examples}
\end{figure*}

\mbox{}\\
Table ~\ref{table:Comparison of Median Cosine Similarities Between Embedding Examples}  and Figure ~\ref{fig:t-SNE Visualization of Graph Embedding Examples}  display the graph embedding results. Embeddings for each example were adjusted using standard scaling. Table ~\ref{table:Comparison of Median Cosine Similarities Between Embedding Examples} illustrates the changes in median cosine similarity among entities within the same class and between entities in different classes as \textit{Class Granularity} varies. As \textit{Class Granularity} increases, there is a tendency for entities belonging to the newly added classes to exhibit higher similarity with other entities within the same class. Particularly when comparing Example1 and Example3, the similarity among entities within the same class increased. Figure ~\ref{fig:t-SNE Visualization of Graph Embedding Examples} provides a t-SNE visualization of the embedding results based on the ontology. It's apparent that as the ontology becomes more intricate, entities tend to cluster more effectively according to their classes.

\begin{table*}
    \centering
    \caption{Example Ontologies for Graph Embedding Experiments}
    \scriptsize 
    \renewcommand{\arraystretch}{2} 
    \begin{tabular}{|p{0.1\textwidth}|p{0.1\textwidth}|p{0.4\textwidth}|p{0.4\textwidth}|}
    \hline
    & \textbf{Example 1} & \textbf{Example 2} & \textbf{Example 3} \\
    \hline
    ontology & 
    \multirow{5}{*}{
        \begin{forest}
            for tree={l=8mm, s=3mm, anchor=north} 
            [Thing, baseline
                [Person, name=person1]
            ]
        \end{forest}
    } &
    \multirow{5}{*}{
        \begin{forest}
            for tree={l=3mm, s=3mm, anchor=north} 
            [Thing, baseline
                [Person, name=person2
                    [Artist]
                    [Athlete]
                    [Entrepreneur]
                    [Politician]
                ]
            ]
        \end{forest}
    } &
    \multirow{5}{*}{
        \begin{forest}
            for tree={l=5mm, s=2mm, anchor=north} 
            [Thing, baseline
                [Person, name=person3
                    [Artist
                        [Actor]
                        [Musician]
                        [Author]
                    ]
                    [Athlete]
                    [Entrepreneur]
                    [Politician]
                ]
            ]
        \end{forest}
    } \\
    & & & \\
    & & & \\
    & & & \\
    & & & \\
    \hline
     class granularity & 0.0813 & 0.3634 & 0.3691 \\
    \hline
    \end{tabular}
    \label{tab:example_ontologies}
\end{table*}

\begin{table*}[t]
\centering
\caption{Comparison of Median Cosine Similarities Between Embedding Examples}
\captionsetup{justification=centerlast}
\begin{subtable}[t]{1\textwidth}
\centering
\small
\begin{tabular}{|c|c|c|c|c|c|c|}
\hline
\multicolumn{1}{|c|}{\textbf{}} & \multicolumn{1}{c|}{\textbf{Actor}} & \multicolumn{1}{c|}{\textbf{Musician}} & \multicolumn{1}{c|}{\textbf{Writer}} & \multicolumn{1}{c|}{\textbf{Athlete}} & \multicolumn{1}{c|}{\textbf{Entrepreneur}} & \multicolumn{1}{c|}{\textbf{Politician}} \\
\hline
\textbf{Actor} & 0.0195 & -0.0272 & -0.0142 & -0.005 & 0.0047 & -0.0224 \\
\hline
\textbf{Musician} & & 0.0513 & -0.024 & -0.0056 & -0.023 & 0.0042 \\
\hline
\textbf{Writer} & & & 0.1329 & 0.0021 & -0.0152 & 0.014 \\
\hline
\textbf{Athlete} & & & & 0.0502 & -0.0213 & 0.0065 \\
\hline
\textbf{Entrepreneur} & & & & & 0.0627 & -0.0267 \\
\hline
\textbf{Politician} & & & & & & 0.0485 \\
\hline
\end{tabular}
\caption{Example 1 - Median Cosine Similarities}
\label{table:example1}
\end{subtable}
\begin{subtable}[t]{1\textwidth}
\centering
\small
\renewcommand{\arraystretch}{0.9}
\begin{tabular}{|c|c|c|c|c|c|c|}
\hline
\multicolumn{1}{|c|}{\textbf{}} & \multicolumn{1}{c|}{\textbf{Actor}} & \multicolumn{1}{c|}{\textbf{Musician}} & \multicolumn{1}{c|}{\textbf{Writer}} & \multicolumn{1}{c|}{\textbf{Athlete}} & \multicolumn{1}{c|}{\textbf{Entrepreneur}} & \multicolumn{1}{c|}{\textbf{Politician}} \\
\hline
\textbf{Actor} & 0.0315 (\textcolor{blue}{+}) & -0.0417 (\textcolor{red}{-}) & -0.0064 (\textcolor{blue}{+}) & -0.0228 (\textcolor{red}{-}) & 0.0229 (\textcolor{blue}{+}) & -0.0398 (\textcolor{red}{-}) \\
\hline
\textbf{Musician} & & 0.1084 (\textcolor{blue}{+}) & -0.0441 (\textcolor{red}{-}) & -0.0134 (\textcolor{red}{-}) & -0.0555 (\textcolor{red}{-}) & -0.0073 (\textcolor{red}{-}) \\
\hline
\textbf{Writer} & & & 0.1213 (\textcolor{red}{-}) & 0.0139 (\textcolor{blue}{+}) & -0.0148 (\textcolor{blue}{+}) & 0.0269 (\textcolor{blue}{+}) \\
\hline
\textbf{Athlete} & & & & 0.1289 (\textcolor{blue}{+}) & -0.0287 (\textcolor{red}{-}) & 0.0335 (\textcolor{blue}{+}) \\
\hline
\textbf{Entrepreneur} & & & & & 0.0933 (\textcolor{blue}{+}) & -0.0394 (\textcolor{red}{-}) \\
\hline
\textbf{Politician} & & & & & & 0.1087 (\textcolor{blue}{+}) \\
\hline
\end{tabular}
\caption{Example 2 - Median Cosine Similarities (Compared to Example 1)}
\label{table:example2}
\end{subtable}

\begin{subtable}[t]{1\textwidth}
\centering
\small
\renewcommand{\arraystretch}{0.9}
\begin{tabular}{|c|c|c|c|c|c|c|}
\hline
\multicolumn{1}{|c|}{\textbf{}} & \multicolumn{1}{c|}{\textbf{Actor}} & \multicolumn{1}{c|}{\textbf{Musician}} & \multicolumn{1}{c|}{\textbf{Writer}} & \multicolumn{1}{c|}{\textbf{Athlete}} & \multicolumn{1}{c|}{\textbf{Entrepreneur}} & \multicolumn{1}{c|}{\textbf{Politician}} \\
\hline
\textbf{Actor} & 0.0399 (\textcolor{blue}{+}) & -0.0609 (\textcolor{red}{-}) & -0.0121 (\textcolor{red}{-}) & -0.0087 (\textcolor{blue}{+}) & 0.0168 (\textcolor{blue}{+}) & -0.0348 (\textcolor{blue}{+}) \\
\hline
\textbf{Musician} & & 0.1346 (\textcolor{blue}{+}) & -0.0352 (\textcolor{blue}{+}) & -0.0031 (\textcolor{blue}{+}) & -0.0462 (\textcolor{blue}{+}) & -0.0128 (\textcolor{red}{-}) \\
\hline
\textbf{Writer} & & & 0.1833 (\textcolor{blue}{+}) & -0.0148 (\textcolor{red}{-}) & -0.0112 (\textcolor{blue}{+}) & 0.0184 (\textcolor{red}{-}) \\
\hline
\textbf{Athlete} & & & & 0.0953 (\textcolor{blue}{-}) & -0.0308 (\textcolor{red}{-}) & 0.0121 (\textcolor{red}{+}) \\
\hline
\textbf{Entrepreneur} & & & & & 0.0952 (\textcolor{blue}{+}) & -0.038 (\textcolor{blue}{+}) \\
\hline
\textbf{Politician} & & & & & & 0.1168 (\textcolor{blue}{+}) \\
\hline
\end{tabular}
\caption{Example 3 - Median Cosine Similarities (Compared to Example 2)}
\label{table:example3}
\end{subtable}
\label{table:Comparison of Median Cosine Similarities Between Embedding Examples}
\end{table*}

\subsection{Knowledge Graph Resoning and Question Answering}
Knowledge graph reasoning is the task of utilizing existing information to discover unknown relationships within the graph. This task is often performed using rule-based approaches with ontology or graph embedding-based link prediction methods. Researches like \citet{10.1145/3308558.3313612,10.1007/978-3-030-77385-4_22,kaoudi2022towards} have shown that combining ontology-based reasoning and embedding-based link prediction can enhance the performance of reasoning. 
Highe \textit{Class Granularity} can have a positive impact on ontology-based reasoning.

Ontology-based reasoning involves creating rules based on the information in the ontology to generate new triples. Having predicates divided into specific classes allows for the generation of rules as finely as needed. For example, someone may want to create rules from the ontology like the following. 

\begin{align*}
\scriptsize
\forall x, y, z \in E: & \\
& \begin{array}{l}
    (x, \text{ownerOf}, y), \\
    (y, \text{industry}, z) \rightarrow \\
    (x, \text{relatedIndustry}, z)
\end{array}
\end{align*}

However, applying such rules to all entities of every class can lead to many unnecessary instances. In this case, if the \text{Person} class is classified as a subclass of \textit{Entrepreneur}, then the rule can be selectively applied to relevant \textit{Entrepreneurs}, making it more useful. This way, the rules can be more targeted and efficient in their application.

The ability to add finely-grained reasoning rules can also be beneficial for Knowledge Base Question Answering(KBQA) tasks. Even in situations where there are no triples in the knowledge graph, having axioms created through ontology's class and predicates can provide answers. This approach is utilized in many KBQA researches\cite{li2019ontology,abdi2018qapd,manna2017cookingqa,fawei2019semi}. For instance, when answering a question like \textit{'Who is an athlete born in America?'} if people whose occupation is athlete only have triples such as \textit{'A-occupation-volleyball player'} and \textit{'B-occupation-baseball player'} without ontology's hierarchical information between classes, it would be impossible to identify these entities as athletes. Alternatively, you would need to add triples like \textit{'A-occupation-athlete'} to all entities who are athletes. However, if the ontology is structured with \textit{Athlete} as a sperclass of \textit{Volleyball Player} and instances are categorized accordingly, defining a single rule can facilitate answering the question.

\subsection{Named Entity Disambiguation}
There are researches focusing on matching text to knowledge graphs, such as NED (Named Entity Disambiguation) or entity linking in fields such as medicine \cite{skreta2021automatically,mondal2020medical} , biology \cite{mohan2021low,halioui2018bioinformatic,farazi2018ontology}, chemistry \cite{wang2021chemner,wang2019ontology}. If there are multiple knowledge graphs in the same domain that can link entities with text, performance of NED can be anticipated the by referring to the value of \textit{Class Granularity}. A high \textit{Class Granularity} indicates that entities are divided into various subfields, and they possess the distinct characteristics of those subfields in the form of triples. For example, in the example previously mentioned in Table ~\ref{tab:Class Granularity Calculation Example Ontologies}, when \textit{Class Granularity} is high, \textit{Music Song}'s instances have more attributes related to songs, and \textit{Movie}'s instances have more attributes related to movies. Therefore, when there is an ambiguity word like \textit{'Love'}, in a high \textit{Class Granularity} setting, the movie 'Love' would have movie-related attributes, and the song 'Love' would have song-related attributes, making disambiguation easier.

\begin{table*}
\centering
\scriptsize 
\caption{Metric Comparison of LOD and Raftel}
\label{Metric Comparison of LOD and Raftel}
\begin{tabular}{
  |p{0.1\textwidth}|p{0.1\textwidth}|p{0.12\textwidth}|p{0.12\textwidth}|p{0.12\textwidth}|p{0.15\textwidth}|p{0.1\textwidth}|
}
\hline
         & Classes & Predicates & Instances & Triples & Avg. Predicates per Class & Granularity \\
\hline
DBpedia  & 472               & 33,457                & 6,570,879             & 60,451,631      & 599                            & 0.0904            \\
\hline
YAGO     & 111               & 133                  & 64,611,470                & 461,321,787     & 24                             & 0.1708            \\
\hline
Freebase & 7425              & 769,935              & 115,755,706             & 961,192,099     & 278                            & 0.3964            \\
\hline
Raftel   & 287               & 1079                 & 28,652,479               & 298,359,151     & 132                            & 0.1400            \\            
\hline
\end{tabular}
\end{table*}

\section{Class Granularity of Linked Open Data}
Table ~\ref{Metric Comparison of LOD and Raftel} provides basic statistics and class granularity calculation results for \textit{Freebase}, \textit{YAGO}, and \textit{DBpedia}. This comparison shows how richly the ontologies are defined in each Linked Open Data (LOD) source and how well they represent the knowledge graph with RDF triples, addressing aspects that have not been extensively explored in previous research on LOD comparison analysis. The process of obtaining RDF triples and ontologies for each LOD is detailed in the Appendix ~\ref{app:The process of extracting LOD and their ontologies}. For each dataset, we filtered only the entities that are instances of classes defined in the ontology among the RDF triples. In cases where obtaining the ontology that defines predicates was challenging, we extracted predicates used for each class from the triples. In addition to LOD, we also conducted a quantitative comparison for Naver's knowledge graph, Raftel, a knowledge graph constructed by consolidating Wikidata's ontology.

 When analyzing the results from Table ~\ref{Metric Comparison of LOD and Raftel}, looking at \textit{Class Granularity} alongside basic metrics such as the number of classes and predicates allows for a more multidimensional understanding of knowledge graphs. For example, when comparing DBpedia and YAGO, DBpedia has more classes and predicates, but its \textit{Class Granularity} is lower. By only considering the quantity, one might conclude that DBpedia has richer information. However, when taking \textit{Class Granularity} into account, it becomes apparent that, relative to the variety of predicates, DBpedia actually has a lower proportion of entities with those predicates. On the other hand, YAGO has fewer predicates in use, but its \textit{Class Granularity} is higher, with an average of 2.779 entities per class having distinct predicates, which is higher than DBpedia. Having a high \textit{Class Granularity} doesn't necessarily imply superiority, but it does provide a way to gauge how well classes possess distinct characteristics beyond just their quantity, which is often hard to evaluate solely based on the number of classes and predicates. 

For the class \textit{Book}, DBpedia has a wider variety of \textit{distinct predicates} such as \textit{shortSummary}, \textit{theme}, and \textit{screenplay}, with each of these predicates having instance ratios of 0.00004, 0.00004, and 0.0001, respectively within the class \textit{Book}. In contrast, YAGO has fewer predicates for the class \textit{Book}, but it has higher instance ratios for predicates like \textit{about} (0.17) and \textit{contentLocation}(0.016), indicating that these predicates are more commonly associated with instances of the class \textit{Book} in YAGO. Specifically, for the common property \textit{publishedDate}, DBpedia has an instance ratio of 0.00024, while YAGO has a higher instance ratio of 0.43879, indicating that a larger proportion of \textit{Book}'s instances in YAGO have this property.

\section{Conclusion and Future Work}

In this study, we introduced \textit{Class Granularity} as a metric for assessing the extent to which classes are finely detailed in knowledge graphs and ontologies. We also highlighted scenarios where this metric can be beneficial. However, it's important to note that higher \textit{Class Granularity} is not always better in every context. For instance, in cases where an ontology is designed with predefined classes and predicates to accommodate future data, even if there are no actual instances, the \textit{Class Granularity} may be relatively low but can enhance flexibility when new data is added.

\textit{Class Granularity} is significant because it quantifies the "structural richness" in knowledge graph quality, which has not received as much attention in previous research. Future studies could explore the correlations between Class Granularity and other dimensions of quality metrics to gain a deeper understanding of how various quality indicators interact in knowledge graphs. Additionally, research could investigate a broader range of knowledge graph applications that may be influenced by changes in \textit{Class Granularity}.

\bibliographystyle{ACM-Reference-Format}
\bibliography{acl2023}


\begin{thebibliography}{42}


\ifx \showCODEN    \undefined \def \showCODEN     #1{\unskip}     \fi
\ifx \showDOI      \undefined \def \showDOI       #1{#1}\fi
\ifx \showISBNx    \undefined \def \showISBNx     #1{\unskip}     \fi
\ifx \showISBNxiii \undefined \def \showISBNxiii  #1{\unskip}     \fi
\ifx \showISSN     \undefined \def \showISSN      #1{\unskip}     \fi
\ifx \showLCCN     \undefined \def \showLCCN      #1{\unskip}     \fi
\ifx \shownote     \undefined \def \shownote      #1{#1}          \fi
\ifx \showarticletitle \undefined \def \showarticletitle #1{#1}   \fi
\ifx \showURL      \undefined \def \showURL       {\relax}        \fi
\providecommand\bibfield[2]{#2}
\providecommand\bibinfo[2]{#2}
\providecommand\natexlab[1]{#1}
\providecommand\showeprint[2][]{arXiv:#2}

\bibitem[Abdi et~al\mbox{.}(2018)]%
        {abdi2018qapd}
\bibfield{author}{\bibinfo{person}{Asad Abdi}, \bibinfo{person}{Norisma Idris}, {and} \bibinfo{person}{Zahrah Ahmad}.} \bibinfo{year}{2018}\natexlab{}.
\newblock \showarticletitle{QAPD: an ontology-based question answering system in the physics domain}.
\newblock \bibinfo{journal}{\emph{Soft Computing}}  \bibinfo{volume}{22} (\bibinfo{year}{2018}), \bibinfo{pages}{213--230}.
\newblock


\bibitem[Alani and Brewster(2006)]%
        {alani2006metrics}
\bibfield{author}{\bibinfo{person}{Harith Alani} {and} \bibinfo{person}{Christopher Brewster}.} \bibinfo{year}{2006}\natexlab{}.
\newblock \showarticletitle{Metrics for ranking ontologies}.
\newblock  (\bibinfo{year}{2006}).
\newblock


\bibitem[Allen and Yen(2001)]%
        {allen2001introduction}
\bibfield{author}{\bibinfo{person}{Mary~J Allen} {and} \bibinfo{person}{Wendy~M Yen}.} \bibinfo{year}{2001}\natexlab{}.
\newblock \bibinfo{booktitle}{\emph{Introduction to measurement theory}}.
\newblock \bibinfo{publisher}{Waveland Press}.
\newblock


\bibitem[Bordes et~al\mbox{.}(2013)]%
        {bordes2013translating}
\bibfield{author}{\bibinfo{person}{Antoine Bordes}, \bibinfo{person}{Nicolas Usunier}, \bibinfo{person}{Alberto Garcia-Duran}, \bibinfo{person}{Jason Weston}, {and} \bibinfo{person}{Oksana Yakhnenko}.} \bibinfo{year}{2013}\natexlab{}.
\newblock \showarticletitle{Translating embeddings for modeling multi-relational data}.
\newblock \bibinfo{journal}{\emph{Advances in neural information processing systems}}  \bibinfo{volume}{26} (\bibinfo{year}{2013}).
\newblock


\bibitem[Brank et~al\mbox{.}(2005)]%
        {brank2005survey}
\bibfield{author}{\bibinfo{person}{Janez Brank}, \bibinfo{person}{Marko Grobelnik}, {and} \bibinfo{person}{Dunja Mladenic}.} \bibinfo{year}{2005}\natexlab{}.
\newblock \showarticletitle{A survey of ontology evaluation techniques}. In \bibinfo{booktitle}{\emph{Proceedings of the conference on data mining and data warehouses (SiKDD 2005)}}. Citeseer, \bibinfo{pages}{166--170}.
\newblock


\bibitem[Debattista et~al\mbox{.}(2016)]%
        {debattista2016luzzu}
\bibfield{author}{\bibinfo{person}{Jeremy Debattista}, \bibinfo{person}{S{\"o}ren Auer}, {and} \bibinfo{person}{Christoph Lange}.} \bibinfo{year}{2016}\natexlab{}.
\newblock \showarticletitle{Luzzu—a methodology and framework for linked data quality assessment}.
\newblock \bibinfo{journal}{\emph{Journal of Data and Information Quality (JDIQ)}} \bibinfo{volume}{8}, \bibinfo{number}{1} (\bibinfo{year}{2016}), \bibinfo{pages}{1--32}.
\newblock


\bibitem[Debattista et~al\mbox{.}(2018)]%
        {debattista2018evaluating}
\bibfield{author}{\bibinfo{person}{Jeremy Debattista}, \bibinfo{person}{Christoph Lange}, \bibinfo{person}{S{\"o}ren Auer}, {and} \bibinfo{person}{Dominic Cortis}.} \bibinfo{year}{2018}\natexlab{}.
\newblock \showarticletitle{Evaluating the quality of the LOD cloud: An empirical investigation}.
\newblock \bibinfo{journal}{\emph{Semantic Web}} \bibinfo{volume}{9}, \bibinfo{number}{6} (\bibinfo{year}{2018}), \bibinfo{pages}{859--901}.
\newblock


\bibitem[Diaz et~al\mbox{.}(2018)]%
        {diaz2018embeds}
\bibfield{author}{\bibinfo{person}{Gonzalo~I Diaz}, \bibinfo{person}{Achille Fokoue}, \bibinfo{person}{Mohammad Sadoghi}, {et~al\mbox{.}}} \bibinfo{year}{2018}\natexlab{}.
\newblock \showarticletitle{EmbedS: Scalable, Ontology-aware Graph Embeddings.}. In \bibinfo{booktitle}{\emph{EDBT}}. \bibinfo{pages}{433--436}.
\newblock


\bibitem[Farazi et~al\mbox{.}(2018)]%
        {farazi2018ontology}
\bibfield{author}{\bibinfo{person}{Feroz Farazi}, \bibinfo{person}{Craig Chapman}, \bibinfo{person}{Pathmeswaran Raju}, {and} \bibinfo{person}{Lynsey Melville}.} \bibinfo{year}{2018}\natexlab{}.
\newblock \showarticletitle{Ontology-based faceted semantic search with automatic sense disambiguation for bioenergy domain}.
\newblock \bibinfo{journal}{\emph{International Journal of Big Data Intelligence}} \bibinfo{volume}{5}, \bibinfo{number}{1-2} (\bibinfo{year}{2018}), \bibinfo{pages}{62--72}.
\newblock


\bibitem[F{\"a}rber and Rettinger(2018)]%
        {farber2018knowledge}
\bibfield{author}{\bibinfo{person}{Michael F{\"a}rber} {and} \bibinfo{person}{Achim Rettinger}.} \bibinfo{year}{2018}\natexlab{}.
\newblock \showarticletitle{Which knowledge graph is best for me?}
\newblock \bibinfo{journal}{\emph{arXiv preprint arXiv:1809.11099}} (\bibinfo{year}{2018}).
\newblock


\bibitem[Fawei et~al\mbox{.}(2019)]%
        {fawei2019semi}
\bibfield{author}{\bibinfo{person}{Biralatei Fawei}, \bibinfo{person}{Jeff~Z Pan}, \bibinfo{person}{Martin Kollingbaum}, {and} \bibinfo{person}{Adam~Z Wyner}.} \bibinfo{year}{2019}\natexlab{}.
\newblock \showarticletitle{A semi-automated ontology construction for legal question answering}.
\newblock \bibinfo{journal}{\emph{New Generation Computing}}  \bibinfo{volume}{37} (\bibinfo{year}{2019}), \bibinfo{pages}{453--478}.
\newblock


\bibitem[Fern{\'a}ndez et~al\mbox{.}(2009)]%
        {10.1007/978-3-642-10871-6_5}
\bibfield{author}{\bibinfo{person}{Miriam Fern{\'a}ndez}, \bibinfo{person}{Chwhynny Overbeeke}, \bibinfo{person}{Marta Sabou}, {and} \bibinfo{person}{Enrico Motta}.} \bibinfo{year}{2009}\natexlab{}.
\newblock \showarticletitle{What Makes a Good Ontology? A Case-Study in Fine-Grained Knowledge Reuse}. In \bibinfo{booktitle}{\emph{The Semantic Web}}, \bibfield{editor}{\bibinfo{person}{Asunci{\'o}n G{\'o}mez-P{\'e}rez}, \bibinfo{person}{Yong Yu}, {and} \bibinfo{person}{Ying Ding}} (Eds.). \bibinfo{publisher}{Springer Berlin Heidelberg}, \bibinfo{address}{Berlin, Heidelberg}, \bibinfo{pages}{61--75}.
\newblock
\showISBNx{978-3-642-10871-6}


\bibitem[Gangemi et~al\mbox{.}(2006)]%
        {gangemi2006modelling}
\bibfield{author}{\bibinfo{person}{Aldo Gangemi}, \bibinfo{person}{Carola Catenacci}, \bibinfo{person}{Massimiliano Ciaramita}, {and} \bibinfo{person}{Jos Lehmann}.} \bibinfo{year}{2006}\natexlab{}.
\newblock \showarticletitle{Modelling ontology evaluation and validation}. In \bibinfo{booktitle}{\emph{European semantic web conference}}. Springer, \bibinfo{pages}{140--154}.
\newblock


\bibitem[G{\'o}mez-P{\'e}rez(2004)]%
        {gomez2004ontology}
\bibfield{author}{\bibinfo{person}{Asunci{\'o}n G{\'o}mez-P{\'e}rez}.} \bibinfo{year}{2004}\natexlab{}.
\newblock \showarticletitle{Ontology evaluation}.
\newblock In \bibinfo{booktitle}{\emph{Handbook on ontologies}}. \bibinfo{publisher}{Springer}, \bibinfo{pages}{251--273}.
\newblock


\bibitem[Gruber(1993)]%
        {gruber1993ontology}
\bibfield{author}{\bibinfo{person}{Tom Gruber}.} \bibinfo{year}{1993}\natexlab{}.
\newblock \bibinfo{title}{What is an Ontology}.
\newblock
\newblock


\bibitem[Halioui et~al\mbox{.}(2018)]%
        {halioui2018bioinformatic}
\bibfield{author}{\bibinfo{person}{Ahmed Halioui}, \bibinfo{person}{Petko Valtchev}, {and} \bibinfo{person}{Abdoulaye~Banire Diallo}.} \bibinfo{year}{2018}\natexlab{}.
\newblock \showarticletitle{Bioinformatic workflow extraction from scientific texts based on word sense disambiguation}.
\newblock \bibinfo{journal}{\emph{IEEE/ACM transactions on computational biology and bioinformatics}} \bibinfo{volume}{15}, \bibinfo{number}{6} (\bibinfo{year}{2018}), \bibinfo{pages}{1979--1990}.
\newblock


\bibitem[Heinrich et~al\mbox{.}(2018)]%
        {heinrich2018requirements}
\bibfield{author}{\bibinfo{person}{Bernd Heinrich}, \bibinfo{person}{Diana Hristova}, \bibinfo{person}{Mathias Klier}, \bibinfo{person}{Alexander Schiller}, {and} \bibinfo{person}{Michael Szubartowicz}.} \bibinfo{year}{2018}\natexlab{}.
\newblock \showarticletitle{Requirements for data quality metrics}.
\newblock \bibinfo{journal}{\emph{Journal of Data and Information Quality (JDIQ)}} \bibinfo{volume}{9}, \bibinfo{number}{2} (\bibinfo{year}{2018}), \bibinfo{pages}{1--32}.
\newblock


\bibitem[Heist et~al\mbox{.}(2020)]%
        {heist2020knowledge}
\bibfield{author}{\bibinfo{person}{Nicolas Heist}, \bibinfo{person}{Sven Hertling}, \bibinfo{person}{Daniel Ringler}, {and} \bibinfo{person}{Heiko Paulheim}.} \bibinfo{year}{2020}\natexlab{}.
\newblock \showarticletitle{Knowledge Graphs on the Web-An Overview.}
\newblock \bibinfo{journal}{\emph{Knowledge Graphs for eXplainable Artificial Intelligence}} (\bibinfo{year}{2020}), \bibinfo{pages}{3--22}.
\newblock


\bibitem[Jain et~al\mbox{.}(2021)]%
        {jain2021improving}
\bibfield{author}{\bibinfo{person}{Nitisha Jain}, \bibinfo{person}{Trung-Kien Tran}, \bibinfo{person}{Mohamed~H Gad-Elrab}, {and} \bibinfo{person}{Daria Stepanova}.} \bibinfo{year}{2021}\natexlab{}.
\newblock \showarticletitle{Improving knowledge graph embeddings with ontological reasoning}. In \bibinfo{booktitle}{\emph{International Semantic Web Conference}}. Springer, \bibinfo{pages}{410--426}.
\newblock


\bibitem[Kaoudi et~al\mbox{.}(2022)]%
        {kaoudi2022towards}
\bibfield{author}{\bibinfo{person}{Zoi Kaoudi}, \bibinfo{person}{Abelardo Carlos~Martinez Lorenzo}, {and} \bibinfo{person}{Volker Markl}.} \bibinfo{year}{2022}\natexlab{}.
\newblock \showarticletitle{Towards loosely-coupling knowledge graph embeddings and ontology-based reasoning}.
\newblock \bibinfo{journal}{\emph{arXiv preprint arXiv:2202.03173}} (\bibinfo{year}{2022}).
\newblock


\bibitem[Lan et~al\mbox{.}(2021)]%
        {lan2021survey}
\bibfield{author}{\bibinfo{person}{Yunshi Lan}, \bibinfo{person}{Gaole He}, \bibinfo{person}{Jinhao Jiang}, \bibinfo{person}{Jing Jiang}, \bibinfo{person}{Wayne~Xin Zhao}, {and} \bibinfo{person}{Ji-Rong Wen}.} \bibinfo{year}{2021}\natexlab{}.
\newblock \showarticletitle{A survey on complex knowledge base question answering: Methods, challenges and solutions}.
\newblock \bibinfo{journal}{\emph{arXiv preprint arXiv:2105.11644}} (\bibinfo{year}{2021}).
\newblock


\bibitem[Li et~al\mbox{.}(2019)]%
        {li2019ontology}
\bibfield{author}{\bibinfo{person}{Wenjia Li}, \bibinfo{person}{Liang Wu}, \bibinfo{person}{Zhong Xie}, \bibinfo{person}{Liufeng Tao}, \bibinfo{person}{Kuanmao Zou}, \bibinfo{person}{Fengdan Li}, {and} \bibinfo{person}{Jinli Miao}.} \bibinfo{year}{2019}\natexlab{}.
\newblock \showarticletitle{Ontology-based question understanding with the constraint of Spatio-temporal geological knowledge}.
\newblock \bibinfo{journal}{\emph{Earth Science Informatics}}  \bibinfo{volume}{12} (\bibinfo{year}{2019}), \bibinfo{pages}{599--613}.
\newblock


\bibitem[Li et~al\mbox{.}(2023)]%
        {li2023transo}
\bibfield{author}{\bibinfo{person}{Zhao Li}, \bibinfo{person}{Xin Liu}, \bibinfo{person}{Xin Wang}, \bibinfo{person}{Pengkai Liu}, {and} \bibinfo{person}{Yuxin Shen}.} \bibinfo{year}{2023}\natexlab{}.
\newblock \showarticletitle{Transo: a knowledge-driven representation learning method with ontology information constraints}.
\newblock \bibinfo{journal}{\emph{World Wide Web}} \bibinfo{volume}{26}, \bibinfo{number}{1} (\bibinfo{year}{2023}), \bibinfo{pages}{297--319}.
\newblock


\bibitem[Liu et~al\mbox{.}(2021)]%
        {10.1007/978-3-030-77385-4_22}
\bibfield{author}{\bibinfo{person}{Yushan Liu}, \bibinfo{person}{Marcel Hildebrandt}, \bibinfo{person}{Mitchell Joblin}, \bibinfo{person}{Martin Ringsquandl}, \bibinfo{person}{Rime Raissouni}, {and} \bibinfo{person}{Volker Tresp}.} \bibinfo{year}{2021}\natexlab{}.
\newblock \showarticletitle{Neural Multi-hop Reasoning with Logical Rules on Biomedical Knowledge Graphs}. In \bibinfo{booktitle}{\emph{The Semantic Web}}, \bibfield{editor}{\bibinfo{person}{Ruben Verborgh}, \bibinfo{person}{Katja Hose}, \bibinfo{person}{Heiko Paulheim}, \bibinfo{person}{Pierre-Antoine Champin}, \bibinfo{person}{Maria Maleshkova}, \bibinfo{person}{Oscar Corcho}, \bibinfo{person}{Petar Ristoski}, {and} \bibinfo{person}{Mehwish Alam}} (Eds.). \bibinfo{publisher}{Springer International Publishing}, \bibinfo{address}{Cham}, \bibinfo{pages}{375--391}.
\newblock
\showISBNx{978-3-030-77385-4}


\bibitem[Lourdusamy and John(2018)]%
        {lourdusamy2018review}
\bibfield{author}{\bibinfo{person}{Ravi Lourdusamy} {and} \bibinfo{person}{Antony John}.} \bibinfo{year}{2018}\natexlab{}.
\newblock \showarticletitle{A review on metrics for ontology evaluation}. In \bibinfo{booktitle}{\emph{2018 2nd International conference on inventive systems and control (ICISC)}}. IEEE, \bibinfo{pages}{1415--1421}.
\newblock


\bibitem[Lozano-Tello and G{\'o}mez-P{\'e}rez(2004)]%
        {lozano2004ontometric}
\bibfield{author}{\bibinfo{person}{Adolfo Lozano-Tello} {and} \bibinfo{person}{Asunci{\'o}n G{\'o}mez-P{\'e}rez}.} \bibinfo{year}{2004}\natexlab{}.
\newblock \showarticletitle{Ontometric: A method to choose the appropriate ontology}.
\newblock \bibinfo{journal}{\emph{Journal of Database Management (JDM)}} \bibinfo{volume}{15}, \bibinfo{number}{2} (\bibinfo{year}{2004}), \bibinfo{pages}{1--18}.
\newblock


\bibitem[Manna et~al\mbox{.}(2017)]%
        {manna2017cookingqa}
\bibfield{author}{\bibinfo{person}{Riyanka Manna}, \bibinfo{person}{Partha Pakray}, \bibinfo{person}{Somnath Banerjee}, \bibinfo{person}{Dipankar Das}, {and} \bibinfo{person}{Alexander Gelbukh}.} \bibinfo{year}{2017}\natexlab{}.
\newblock \showarticletitle{CookingQA: A question answering system based on cooking ontology}. In \bibinfo{booktitle}{\emph{Advances in Computational Intelligence: 15th Mexican International Conference on Artificial Intelligence, MICAI 2016, Canc{\'u}n, Mexico, October 23--28, 2016, Proceedings, Part I 15}}. Springer, \bibinfo{pages}{67--78}.
\newblock


\bibitem[Mohan et~al\mbox{.}(2021)]%
        {mohan2021low}
\bibfield{author}{\bibinfo{person}{Sunil Mohan}, \bibinfo{person}{Rico Angell}, \bibinfo{person}{Nicholas Monath}, {and} \bibinfo{person}{Andrew McCallum}.} \bibinfo{year}{2021}\natexlab{}.
\newblock \showarticletitle{Low resource recognition and linking of biomedical concepts from a large ontology}. In \bibinfo{booktitle}{\emph{Proceedings of the 12th ACM Conference on Bioinformatics, Computational Biology, and Health Informatics}}. \bibinfo{pages}{1--10}.
\newblock


\bibitem[Mondal et~al\mbox{.}(2020)]%
        {mondal2020medical}
\bibfield{author}{\bibinfo{person}{Ishani Mondal}, \bibinfo{person}{Sukannya Purkayastha}, \bibinfo{person}{Sudeshna Sarkar}, \bibinfo{person}{Pawan Goyal}, \bibinfo{person}{Jitesh Pillai}, \bibinfo{person}{Amitava Bhattacharyya}, {and} \bibinfo{person}{Mahanandeeshwar Gattu}.} \bibinfo{year}{2020}\natexlab{}.
\newblock \showarticletitle{Medical entity linking using triplet network}.
\newblock \bibinfo{journal}{\emph{arXiv preprint arXiv:2012.11164}} (\bibinfo{year}{2020}).
\newblock


\bibitem[Obrst et~al\mbox{.}(2007)]%
        {obrst2007evaluation}
\bibfield{author}{\bibinfo{person}{Leo Obrst}, \bibinfo{person}{Werner Ceusters}, \bibinfo{person}{Inderjeet Mani}, \bibinfo{person}{Steve Ray}, {and} \bibinfo{person}{Barry Smith}.} \bibinfo{year}{2007}\natexlab{}.
\newblock \showarticletitle{The evaluation of ontologies: Toward improved semantic interoperability}.
\newblock \bibinfo{journal}{\emph{Semantic web: Revolutionizing knowledge discovery in the life sciences}} (\bibinfo{year}{2007}), \bibinfo{pages}{139--158}.
\newblock


\bibitem[Raad and Cruz(2015)]%
        {raad2015survey}
\bibfield{author}{\bibinfo{person}{Joe Raad} {and} \bibinfo{person}{Christophe Cruz}.} \bibinfo{year}{2015}\natexlab{}.
\newblock \showarticletitle{A survey on ontology evaluation methods}. In \bibinfo{booktitle}{\emph{Proceedings of the International Conference on Knowledge Engineering and Ontology Development, part of the 7th International Joint Conference on Knowledge Discovery, Knowledge Engineering and Knowledge Management}}.
\newblock


\bibitem[Reinanda et~al\mbox{.}(2020)]%
        {reinanda2020knowledge}
\bibfield{author}{\bibinfo{person}{Ridho Reinanda}, \bibinfo{person}{Edgar Meij}, \bibinfo{person}{Maarten de Rijke}, {et~al\mbox{.}}} \bibinfo{year}{2020}\natexlab{}.
\newblock \showarticletitle{Knowledge graphs: An information retrieval perspective}.
\newblock \bibinfo{journal}{\emph{Foundations and Trends{\textregistered} in Information Retrieval}} \bibinfo{volume}{14}, \bibinfo{number}{4} (\bibinfo{year}{2020}), \bibinfo{pages}{289--444}.
\newblock


\bibitem[Ringler and Paulheim(2017)]%
        {ringler2017one}
\bibfield{author}{\bibinfo{person}{Daniel Ringler} {and} \bibinfo{person}{Heiko Paulheim}.} \bibinfo{year}{2017}\natexlab{}.
\newblock \showarticletitle{One knowledge graph to rule them all? Analyzing the differences between DBpedia, YAGO, Wikidata \& co.}. In \bibinfo{booktitle}{\emph{KI 2017: Advances in Artificial Intelligence: 40th Annual German Conference on AI, Dortmund, Germany, September 25--29, 2017, Proceedings 40}}. Springer, \bibinfo{pages}{366--372}.
\newblock


\bibitem[Schneider et~al\mbox{.}(2022)]%
        {schneider2022decade}
\bibfield{author}{\bibinfo{person}{Phillip Schneider}, \bibinfo{person}{Tim Schopf}, \bibinfo{person}{Juraj Vladika}, \bibinfo{person}{Mikhail Galkin}, \bibinfo{person}{Elena Simperl}, {and} \bibinfo{person}{Florian Matthes}.} \bibinfo{year}{2022}\natexlab{}.
\newblock \showarticletitle{A decade of knowledge graphs in natural language processing: A survey}.
\newblock \bibinfo{journal}{\emph{arXiv preprint arXiv:2210.00105}} (\bibinfo{year}{2022}).
\newblock


\bibitem[Shao et~al\mbox{.}(2021)]%
        {shao2021survey}
\bibfield{author}{\bibinfo{person}{Bilin Shao}, \bibinfo{person}{Xiaojun Li}, {and} \bibinfo{person}{Genqing Bian}.} \bibinfo{year}{2021}\natexlab{}.
\newblock \showarticletitle{A survey of research hotspots and frontier trends of recommendation systems from the perspective of knowledge graph}.
\newblock \bibinfo{journal}{\emph{Expert Systems with Applications}}  \bibinfo{volume}{165} (\bibinfo{year}{2021}), \bibinfo{pages}{113764}.
\newblock


\bibitem[Skreta et~al\mbox{.}(2021)]%
        {skreta2021automatically}
\bibfield{author}{\bibinfo{person}{Marta Skreta}, \bibinfo{person}{Aryan Arbabi}, \bibinfo{person}{Jixuan Wang}, \bibinfo{person}{Erik Drysdale}, \bibinfo{person}{Jacob Kelly}, \bibinfo{person}{Devin Singh}, {and} \bibinfo{person}{Michael Brudno}.} \bibinfo{year}{2021}\natexlab{}.
\newblock \showarticletitle{Automatically disambiguating medical acronyms with ontology-aware deep learning}.
\newblock \bibinfo{journal}{\emph{Nature communications}} \bibinfo{volume}{12}, \bibinfo{number}{1} (\bibinfo{year}{2021}), \bibinfo{pages}{5319}.
\newblock


\bibitem[Tartir et~al\mbox{.}(2005)]%
        {tartir2005ontoqa}
\bibfield{author}{\bibinfo{person}{Samir Tartir}, \bibinfo{person}{I~Budak Arpinar}, \bibinfo{person}{Michael Moore}, \bibinfo{person}{Amit~P Sheth}, {and} \bibinfo{person}{Boanerges Aleman-Meza}.} \bibinfo{year}{2005}\natexlab{}.
\newblock \showarticletitle{OntoQA: Metric-based ontology quality analysis}.
\newblock  (\bibinfo{year}{2005}).
\newblock


\bibitem[Vrande{\v{c}}i{\'c}(2009)]%
        {vrandevcic2009ontology}
\bibfield{author}{\bibinfo{person}{Denny Vrande{\v{c}}i{\'c}}.} \bibinfo{year}{2009}\natexlab{}.
\newblock \showarticletitle{Ontology evaluation}.
\newblock In \bibinfo{booktitle}{\emph{Handbook on ontologies}}. \bibinfo{publisher}{Springer}, \bibinfo{pages}{293--313}.
\newblock


\bibitem[Wang et~al\mbox{.}(2019)]%
        {wang2019ontology}
\bibfield{author}{\bibinfo{person}{Rong-Lin Wang}, \bibinfo{person}{Stephen Edwards}, {and} \bibinfo{person}{Cataia Ives}.} \bibinfo{year}{2019}\natexlab{}.
\newblock \showarticletitle{Ontology-based semantic mapping of chemical toxicities}.
\newblock \bibinfo{journal}{\emph{Toxicology}}  \bibinfo{volume}{412} (\bibinfo{year}{2019}), \bibinfo{pages}{89--100}.
\newblock


\bibitem[Wang et~al\mbox{.}(2021)]%
        {wang2021chemner}
\bibfield{author}{\bibinfo{person}{Xuan Wang}, \bibinfo{person}{Vivian Hu}, \bibinfo{person}{Xiangchen Song}, \bibinfo{person}{Shweta Garg}, \bibinfo{person}{Jinfeng Xiao}, {and} \bibinfo{person}{Jiawei Han}.} \bibinfo{year}{2021}\natexlab{}.
\newblock \showarticletitle{ChemNER: fine-grained chemistry named entity recognition with ontology-guided distant supervision}. In \bibinfo{booktitle}{\emph{Proceedings of the 2021 Conference on Empirical Methods in Natural Language Processing}}.
\newblock


\bibitem[Zaveri et~al\mbox{.}(2016)]%
        {zaveri2016quality}
\bibfield{author}{\bibinfo{person}{Amrapali Zaveri}, \bibinfo{person}{Anisa Rula}, \bibinfo{person}{Andrea Maurino}, \bibinfo{person}{Ricardo Pietrobon}, \bibinfo{person}{Jens Lehmann}, {and} \bibinfo{person}{S{\"o}ren Auer}.} \bibinfo{year}{2016}\natexlab{}.
\newblock \showarticletitle{Quality assessment for linked data: A survey}.
\newblock \bibinfo{journal}{\emph{Semantic Web}} \bibinfo{volume}{7}, \bibinfo{number}{1} (\bibinfo{year}{2016}), \bibinfo{pages}{63--93}.
\newblock


\bibitem[Zhang et~al\mbox{.}(2019)]%
        {10.1145/3308558.3313612}
\bibfield{author}{\bibinfo{person}{Wen Zhang}, \bibinfo{person}{Bibek Paudel}, \bibinfo{person}{Liang Wang}, \bibinfo{person}{Jiaoyan Chen}, \bibinfo{person}{Hai Zhu}, \bibinfo{person}{Wei Zhang}, \bibinfo{person}{Abraham Bernstein}, {and} \bibinfo{person}{Huajun Chen}.} \bibinfo{year}{2019}\natexlab{}.
\newblock \showarticletitle{Iteratively Learning Embeddings and Rules for Knowledge Graph Reasoning}. In \bibinfo{booktitle}{\emph{The World Wide Web Conference}} (San Francisco, CA, USA) \emph{(\bibinfo{series}{WWW '19})}. \bibinfo{publisher}{Association for Computing Machinery}, \bibinfo{address}{New York, NY, USA}, \bibinfo{pages}{2366–2377}.
\newblock
\showISBNx{9781450366748}
\urldef\tempurl%
\url{https://doi.org/10.1145/3308558.3313612}
\showDOI{\tempurl}


\end{thebibliography}

\appendix

\section{The process of extracting LOD and their ontologies}\label{app:The process of extracting LOD and their ontologies}

This metric was evaluated by 4 Linked Open Data: DBpedia, YAGO, Freebase, and Wikidata.
The characteristics of each Linked Open Data and the way it is stored are different, and we will explain them in detail below.

\subsection{DBpedia}
DBpedia is a well-known project in the Linked Open Data community.
Its inauguration took place in 2007 with the collaboration of the Free University of Berlin and the University of Leipzig.
It is created by automatically extracting structured information contained in Wikipedia, and it builds and manages its own ontology structure.

The most recent data from DBpedia is accessible via \url{https://databus.dbpedia.org/dbpedia/collections/latest-core} and the version "2022.12.01" was used for the calculation.
The ontology data (\url{ontology--DEV_tag=sorted_type=parsed.nt}) and RDF triple data (\url{infobox-properties_lang=en.ttl.bzip2}, \url{instance-types_inference=specific_lang=en.ttl.bzip2}) have been obtained from their respective pathways, with the exception of when the object part is left empty.
Moreover, we have excluded cases where the ontology name is in unicode characters.

\subsection{YAGO}
Developed by Max Planck in 2007, YAGO is generated by extracting information from Wikipedia infobox and WordNet in various languages.
The ontology structure is built on WordNet.

The most recent YAGO data can be accessed via \url{https://yago-knowledge.org/data/yago4/full/2020-02-24} and the version "2020-02-24" was used as stated in the link.
As with DBpedia, the ontology data(\url{yago-wd-schema.nt.gz}) and RDF triple data(\url{yago-wd-facts.nt.gz}, \url{yago-wd-labels.nt.gz}, \url{yago-wd-class.nt.gz}, \url{yago-wd-sameAs.nt.gz}, \url{yago-wd-full-types.nt.gz}, \url{yago-wd-simple-types.nt.gz}) have been obtained with the exception of when the object part is left empty.

\subsection{Freebase}
Freebase was launched by MetaWeb Technologies, Inc. in 2007 and merged into Wikidata by the Wikimedia Foundation and Google in 2015.
Ontology is constructed in a human-readable manner with the structure of ‘domain/class/predicate’.

Although the official site states that the data dump is no longer updated since 2015, the information available from Freebase can be accessed via \url{http://developers.google.com/freebase} and the link for the data is \url{http://commondatastorage.googleapis.com/freebase-public/rdf/freebase-rdf-latest.gz}.

Since the ontology data and RDF triple data were consolidated into a single file, the ontology data was extracted using specific criteria.
These criteria entailed selecting triples where the predicate was \url{http://www.w3.org/1999/02/22-rdf-syntax-ns#type} and the object was either \url{http://www.w3.org/2002/07/owl#FunctionalProperty} or \url{http://www.w3.org/2000/01/rdf-schema#Property}.
RDF triple data containing a corresponding list of instances pertaining to topics were extracted and defined as ontology data.

Furthermore, as there is no hierarchical relationship between classes in Freebase, a virtual root class called "<Thing>" was created in order to add a hierarchical relationship in the ontology data.
Also, as class information for a particular instance was obtained from other knowledge graphs using a predicate known as \url{<http://rdf.freebase.com/ns/type.type.instance>} (e.g. Person - \url{<http://rdf.freebase.com/ns/type.type.instance>} - Michael Jackson), we reversed the subject and object by employing a predicate referred to as \url{<http://rdf.freebase.com/ns/type.object.type>}. We have renamed the \url{<http://www.w3.org/1999/02/22-rdf-syntax-ns#type>} predicate to \url{<http://rdf.freebase.com/ns/type.object.type>} to create a predicate with a matching definition as \url{<http://rdf.freebase.com/ns/type.object.type>}.

\end{document}